\documentclass{article}

\usepackage[numbers]{natbib}
\usepackage[preprint]{neurips_data_2024}

\usepackage[utf8]{inputenc} 
\usepackage[T1]{fontenc}    
\usepackage{hyperref}       
\usepackage{url}            
\usepackage{booktabs}       
\usepackage{amsfonts}       
\usepackage{nicefrac}       
\usepackage{microtype}      
\usepackage{xcolor}         
\usepackage{amsmath}
\usepackage{bm}
\usepackage{graphicx}
\usepackage{subfigure}
\usepackage{caption}

\usepackage{bbm}
\usepackage{multirow}
\usepackage{wrapfig}
\usepackage{arydshln} 
\usepackage{colortbl}
\usepackage{color,xcolor} 
\usepackage{soul}
\usepackage[normalem]{ulem}
\useunder{\uline}{\ul}{}
\def\eg{\emph{e.g.}}
\def\ie{\emph{i.e.}}

\newcommand{\etal}{\textit{et} \textit{al}.\space}

\newcommand{\tool}{GenderBias-\emph{VL}\space}
\newcommand{\toolns}{GenderBias-\emph{VL}}

\usepackage{cleveref}
\Crefname{figure}{Fig.}{Figs.} 
\crefname{figure}{Fig.}{Figs.} 
\crefname{table}{Tab.}{Tabs.}
\Crefname{table}{Tab.}{Tabs.}

\usepackage{pifont}
\usepackage[perpage,symbol*]{footmisc}
\DefineFNsymbols{circled}{{\ding{192}}{\ding{193}}{\ding{194}}
{\ding{195}}{\ding{196}}{\ding{197}}{\ding{198}}{\ding{199}}{\ding{200}}{\ding{201}}}
\setfnsymbol{circled}

\newcommand\myfootnotestyle[1]{\ifcase#1 \or \ding{182}\or \ding{183}\or
\ding{184}\or \ding{185}\or \ding{186}\or \ding{187}%
\or \ding{188}\or \ding{189}\or \ding{190}\or \ding{191}\else *\fi\relax}

\newcommand*{\affaddr}[1]{#1} 
\newcommand*{\affmark}[1][*]{\textsuperscript{#1}}

\title{
GenderBias-\emph{VL}: Benchmarking Gender Bias in Vision Language Models via Counterfactual Probing}

\author{
    Yisong Xiao\affmark[1],
    Aishan Liu\affmark[1],  
    QianJia Cheng\affmark[1], 
    Zhenfei Yin\affmark[2], 
    Siyuan Liang\affmark[3], \\
    \textbf{Jiapeng Li\affmark[1]}, 
    \textbf{Jing Shao\affmark[2]},  
    \textbf{Xianglong Liu\affmark[1]},
    \textbf{Dacheng Tao\affmark[4]}
    \\
    \affaddr{\affmark[1]Beihang University}
    \affaddr{\affmark[2]Shanghai Artificial Intelligence Laboratory} \\
    \affaddr{\affmark[3]{National University of Singapore}}
    \affaddr{\affmark[4]{Nanyang Technological University}}
}

\begin{document}

\maketitle

\begin{abstract}

Large Vision-Language Models (LVLMs) have been widely adopted in various applications; however, they exhibit significant gender biases. Existing benchmarks primarily evaluate gender bias at the demographic group level, neglecting individual fairness, which emphasizes equal treatment of similar individuals. This research gap limits the detection of discriminatory behaviors, as individual fairness offers a more granular examination of biases that group fairness may overlook. For the first time, this paper introduces the \tool benchmark to evaluate occupation-related gender bias in LVLMs using counterfactual visual questions under individual fairness criteria. 
To construct this benchmark, we first utilize text-to-image diffusion models to generate occupation images and their gender counterfactuals.
Subsequently, we generate corresponding textual occupation options by identifying stereotyped occupation pairs with high semantic similarity but opposite gender proportions in real-world statistics. This method enables the creation of large-scale visual question counterfactuals to expose biases in LVLMs, applicable in both multimodal and unimodal contexts through modifying gender attributes in specific modalities. Overall, our \tool benchmark comprises 34,581 visual question counterfactual pairs,  covering 177 occupations. Using our benchmark, we extensively evaluate 15 commonly used open-source LVLMs (\eg, LLaVA) and state-of-the-art commercial APIs, including GPT-4o and Gemini-Pro. Our findings reveal widespread gender biases in existing LVLMs. Our benchmark offers: (1) a comprehensive dataset for occupation-related gender bias evaluation; (2) an up-to-date leaderboard on LVLM biases; and (3) a nuanced understanding of the biases presented by these models. \footnote{The dataset and code are available at the \href{https://genderbiasvl.github.io/}{website}.}

\end{abstract}
\section{Introduction}
LVLMs have witnessed rapid development \cite{liu2024visual,dai2024instructblip,bai2023qwen}, which expands the capabilities of large language models (LLMs) \cite{chiang2023vicuna,brown2020language,touvron2023llama2} by incorporating additional modalities such as images, showcasing remarkable performance in perceiving and reasoning (\eg, GPT-4V \cite{achiam2023gpt}, Gemini \cite{team2023gemini}). Despite the advancements, there remains a lingering social concern regarding the potential social biases caused by LVLMs \cite{schuhmann2022laion,birhane2021multimodal,caliskan2017semantics}. For instance, research has revealed that ChatGPT associates certain occupations with genders \cite{ghosh2023chatgpt}, depicting doctors as male and nurses as female. 
Such bias discriminates against affected population groups and can significantly harm society as these models are extensively deployed in the real world. Therefore, it is crucial to probe and benchmark the social biases exhibited by LVLMs, serving as a necessary initial step to mitigating the risk of discriminatory outcomes.



While considerable benchmarks \cite{luccioni2024stable,hall2024visogender,howard2023probing,agarwal2021evaluating} have been proposed to measure social biases in LVLMs, they primarily focus on group fairness \cite{hardt2016equality,barocas2016big}, which involve comparing the model performance disparity across different demographic groups (\eg, different gender) \cite{lee2023survey}. 
However, individual fairness \cite{dwork2012fairness,galhotra2017fairness,kusner2017counterfactual} that emphasizes equal treatment of similar individuals has been neglected.
This research gap limits the detection of discriminatory behaviors, as individual fairness could provide a more granular examination of discriminatory behavior across diverse contexts \cite{galhotra2017fairness,xiao2023latent} that group fairness may overlook.

To bridge this gap, we introduce the \tool benchmark, the first to evaluate occupation-related gender bias \cite{de2019bias,wan2024male} in LVLMs using counterfactual visual questions under individual fairness criteria. 
For \textit{visual inputs}, we utilize text-to-image diffusion models. Specifically, we employ Stable Diffusion XL \cite{podell2023sdxl} to generate base occupation images and InstructPix2Pix \cite{brooks2023instructpix2pix} for editing, enabling the creation of counterfactual images that differ in gender attribute. After generation, we filter generated images to ensure they are suitable and valid for testing. 
For \textit{textual options}, we identify stereotyped occupation pairs that share high semantic similarity but exhibit opposite gender proportions in the real world. Drawing data from the U.S. Bureau of Labor Statistics (BLS) \cite{USLabor}, we categorize occupations as male- or female-dominated and pair them accordingly. Then, we utilize CLIP \cite{radford2021learning} to assess and retain pairs with high semantic similarity (\eg, male-dominated \texttt{chief executive} and female-dominated \texttt{executive secretary}).
With this method, we construct large-scale visual question counterfactuals to develop \tool and quantify bias in LVLMs by comparing differences in option selection. Besides, \tool enables bias evaluation in both multimodal and unimodal contexts via controlling the modality of gender counterfactuals. Overall, our \tool comprises 34,581 visual question counterfactual pairs and covers 177 occupations.

Leveraging \toolns, we extensively evaluate 15 commonly used open-source LVLMs (\eg, LLaVA \cite{liu2023improved}) and state-of-the-art commercial APIs (\ie, GPT-4o \cite{gpt4o} and Gemini-Pro \cite{team2023gemini}). From the large-scale experiments, we instantiate several insightful observations as \ding{182} bias permeates LVLMs and becomes especially severe in certain occupation pairs (\eg, \texttt{chief executive} and \texttt{executive secretary}); \ding{183} LVLMs tend to exhibit stronger biases as they become more powerful; \ding{184} gender biases in LVLMs may come from the training corpus as these biases closely mirror existing gender imbalances in real-world labor statistics; and \ding{185} visual and language biases in LVLMs are highly aligned due to the LLM-centric nature. Our benchmark offers: \ding{182} a comprehensive dataset for occupation-related gender bias evaluation; \ding{183} an up-to-date leaderboard on LVLM biases; and \ding{184} a nuanced understanding of the biases presented by these models. We will continuously develop this ecosystem for the community.

\section{Related Works}

\textbf{Large Vision Language Models}. The remarkable success of Large Language Models (LLMs) has sparked Large Vision Language Models (LVLMs) to the research hotspot, such as GPT-4V \cite{achiam2023gpt} and Gemini \cite{team2023gemini}. Typically, these LVLMs \cite{liu2024visual,dai2024instructblip,li2023blip,yin2024lamm,bai2023qwen,team2023internlm,yu2023rlhf,sun2023aligning,chen2023shikra,zhang2023internlm} consist of three components: a CLIP image encoder \cite{radford2021learning,cherti2023reproducible,sun2023eva} for handling visual input, a pre-trained LLM \cite{chiang2023vicuna,touvron2023llama,brown2020language,touvron2023llama2} serving as the system's brain, and a modality interface (\eg, MLP \cite{liu2024visual,liu2023improved} and Q-Former \cite{dai2024instructblip,li2023blip}) aligning different modalities. 
Recently, Qwen-VL \cite{bai2023qwen} and InternLM-XComposer \cite{zhang2023internlm} have further enhanced multimodal comprehension by leveraging diverse task datasets.
Despite their impressive performance in downstream applications, LVLMs often exhibit undesirable behaviors concerning robustness, privacy, and other trustworthiness issues  \cite{wang2021dual,liu2019perceptual,liu2020bias,zhang2021interpreting,tang2021robustart,liu2021training,liu2020spatiotemporal,liu2023x,liu2022harnessing,liu2023exploring,guo2023towards,liu2023towards,liu2023pre,xiao2023robustmq}. In particular, there is a significant lack of research dedicated to probing their social biases. Our work aims to fill this gap by providing a dataset for evaluating occupation-related gender bias in LVLMs.


\textbf{Measuring Bias in LVLMs}. 
Previous research has revealed that vision-language models can absorb social biases inherent in the training data that consists of stereotyped internet-scraped image-text pairs, potentially leading to the propagation and amplification of bias during content generation tasks such as image captioning \cite{bhargava2019exposing,hendricks2018women,qiu2023gender} and retrieval \cite{mitchell2020diversity,zhao2017men}.
Several image-text datasets \cite{zhou2022vlstereoset,hall2024visogender,janghorbani2023multimodal,howard2023probing,zhang2022counterfactually,zhao2021understanding} have been proposed to measure social bias in vision-language models. SocialCounterfactuals \cite{howard2023probing} probes intersectional bias in CLIP models via 171k generated counterfactual image-text pairs under retrieval task. Leveraging SocialCounterfactuals, Howard \etal \cite{howard2024uncovering} evaluated the influence of social attributes (\eg, gender and race) on the toxicity and competency-associated words within content generated by LVLMs. PAIRS \cite{fraser2024examining} comprises 40 images depicting 10 visually ambiguous yet stereotyped working scenarios to examine gender and racial biases in four LVLMs. However, its conclusions are constrained by the small dataset size and unrigorous evaluation methods. 
Moreover, these works measure bias by comparing performance disparities across groups, overlooking granular individual discrimination. In contrast, our \tool offers a detailed and comprehensive evaluation of bias in LVLMs under individual fairness criteria. We note that VL-Bias \cite{zhang2022counterfactually} evaluates individual discrimination within Masked Language Models, but it is not suitable for evaluating LVLMs, and its method of using adversarial attacks to introduce counterfactuals is debatable \cite{lee2023survey}.

\begin{figure*}[t]
\centering
\includegraphics[width=0.95\linewidth]{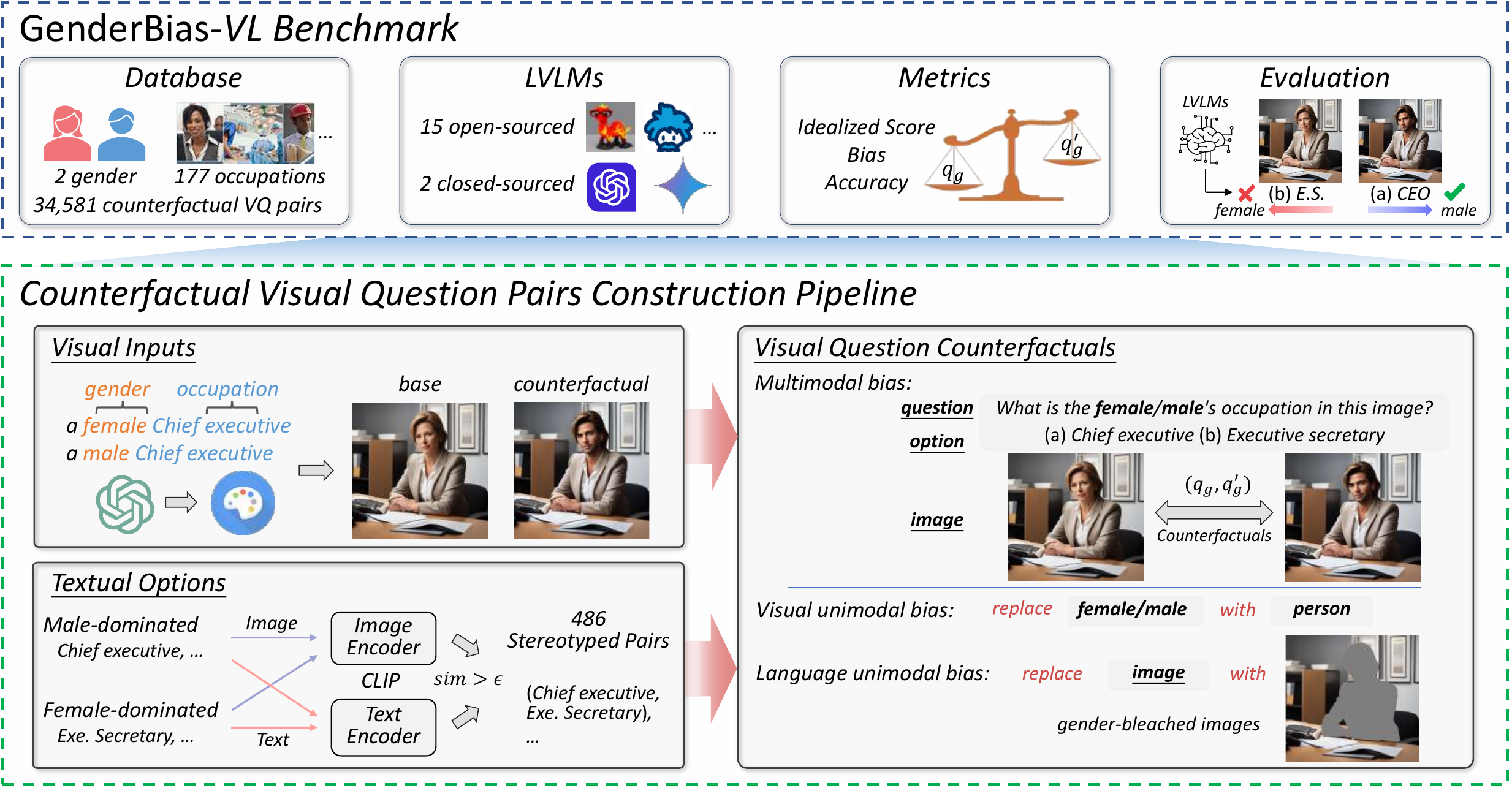}
\caption{Overview of \toolns. we design a construction pipeline to develop \tool, comprising 34,581 visual question counterfactual pairs covering 177 occupations, enabling LVLM bias evaluation in multimodal and unimodal contexts under individual fairness criteria.}
\label{fig:framework}
\end{figure*}

\section{\tool Benchmark}

In this section, we first introduce the terminology used in our paper. We then describe the construction pipeline of our \tool benchmark, detailing the processes of occupation image generation, stereotyped occupation pairs identification, and visual question counterfactuals creation. Finally, we outline the evaluation protocol of \toolns. \cref{fig:framework} provides an overview of our benchmark.

\subsection{Terminology}
\label{sec:terminology}

In this paper, we employ a binary gender framework (\ie, female and male) to facilitate analysis.
We refrain from making claims about gender identification or assignment, recognizing these as deeply personal experiences regardless of appearance or traits \cite{dev2021harms,carothers2013men}. Like previous work \cite{hall2024visogender,howard2023probing,wan2024male}, our study is grounded in the concept of \emph{perceived gender presentation}, which refers to the inference of gender by an external human annotator or model. We acknowledge that the labels used in this study may not fully align with an individual's gender identity. Moreover, we recognize that gender and gender identity are multidimensional notions spanning a spectrum, and acknowledge that misinterpretations within a binary paradigm may occur.

\subsection{Counterfactual Visual Question Pairs Construction Pipeline}
Occupation-related gender bias refers to how changes in gender within inputs affect the LVLM's inference on occupations. Thus, we construct visual questions and their gender counterfactual pairs to evaluate the LVLMs' occupation cognitive ability and bias, respectively. Specifically, LVLMs are tasked with inferring the depicted occupation in the input image from two stereotyped occupation options. Here, a visual question consists of three components: \textbf{\underline{image}}, \textbf{\underline{question}}, and \textbf{\underline{option}}. We now illustrate our counterfactual visual question pairs construction pipeline by explaining each component; with more detailed information (\eg, occupation list) presented in the Appendix.


\textbf{Occupation Image Generation} ({\textbf{\underline{image}}}). We utilize text-to-image diffusion models to generate occupation images and their counterfactuals, following three steps: \emph{prompt creation}, \emph{counterfactual image pairs generation}, and \emph{image filtering}. 

\textbf{Step} \ding{182}: \textit{Prompt Creation.} Given the sets of gender attributes $G$ and occupation names $O$, we denote a subject as $x =$ ``a $\{gender\}$ $\{occ\}$'' (\eg, ``a female Chief executive''), where $gender \in G$ and $occ \in O$. For gender attributes, we employ the terms ``female'' and ``male''; for occupation names, we use a list of 248 occupations from the BLS. Following the guidelines in \cite{SDPromptGuide,SDChatGPTPromptGuide}, we leverage ChatGPT to generate high-quality prompts for text-to-image diffusion models. We iterate through all combinations of gender and occupation values to obtain different subjects, prompting ChatGPT to generate five diverse prompts for each subject.


\textbf{Step} \ding{183}: \textit{Counterfactual Image Pairs Generation.} We employ Stable Diffusion XL \cite{podell2023sdxl}, renowned for its superior photorealism, to generate base occupation images guided by the above prompts. In the realm of image editing, InstructPix2Pix \cite{brooks2023instructpix2pix} achieves strong image consistency pre- and post-editing by training diffusion models on meticulously curated paired images and captions. Consequently, we prompt InstructPix2Pix with a base image (``female''/``male'') and corresponding edit instruction (``turn her into a male''/``turn him into a female''), enabling the creation of counterfactuals that preserve much of the original details while differing in gender. For each subject, we over-generate 100 base images (20 images for each prompt) and their counterfactuals. Additionally, we incorporate a universal negative prompt to steer the images away from undesired artifacts.

\textbf{Step} \ding{184}: \textit{Image Filtering.} To ensure the validity of images for gender bias evaluation, we exclude not-safe-for-work contents \cite{nsfw_detector} and apply filters based on person counts and gender attributes. For person count validation, we use the Grounding DINO \cite{liu2023grounding} object detector to detect bounding boxes of individuals and determine the number of people in each image. Images with person counts other than one are excluded: multiple people may impact model evaluation, while images with no people violate the criterion for human-centric bias evaluation. For gender attributes, we use the CLIP ViT-L-14 \cite{radford2021learning} to classify images as female or male. Both base and counterfactual image pairs must align with the prompted gender during generation to be included; otherwise, they are filtered out.

\textbf{Stereotyped Occupation Pairs Identification} ({\textbf{\underline{option}}}). To expose gender bias in LVLMs, we introduce stereotyped occupation pairs, which share high semantic similarity yet exhibit opposite gender dominance. First, occupations are categorized as male- or female-dominated based on BLS \cite{USLabor} gender proportions, then combined pairwise to generate occupation pairs. 
Then, given $occ_m$ (male-dominated) and $occ_f$ (female-dominated), their base images $\bm{V}_{occ_m}$ and $\bm{V}_{occ_f}$ (100 over-generated images of each occupation), names $\bm{t}_{occ_m}$ and $\bm{t}_{occ_f}$, we utilize CLIP model's vision encoder $E_\text{img}$ and text encoder $E_\text{text}$ to calculate visual and textual similarities as follows:
\begin{equation} 
sim_{img}(occ_m, occ_f) = \langle \bm{I}_{occ_m}, \bm{I}_{occ_f} \rangle, \quad sim_{text}(occ_m, occ_f) = \langle E_\text{text}(\bm{t}_{occ_m}), E_\text{text}(\bm{t}_{occ_f}) \rangle,
\end{equation} 
where $\bm{I} = \frac{1}{|\bm{V}|} \sum\nolimits_{j=1}^{|\bm{V}|}E_\text{img}(\bm{V}_j)$ represents the averaged image embeddings of occupation concept. For all occupation pairs, we collect their similarities, apply min-max normalization within each modality, and compute the final semantic similarity scores as: ${sim} = 0.5 \times [\text{norm}({sim}_{img}) + \text{norm}({sim}_{text})]$.
To select pairs with high semantic similarity, we rank ${sim}$ scores and retain scores higher than $\epsilon$. 
Empirically, $\epsilon$ is set as the lowest score among five commonly identified stereotyped occupation pairs \cite{fraser2024examining}. Note that an occupation may share high similarities with many others, we further restrict each occupation to ten pairings to prioritize the most relevant matches. Finally, we identify 486 stereotyped occupation pairs (covering 177 unique occupations).

\textbf{Visual Question Counterfactuals Creation}. Based on the above {image} and {option} components, we can devise \textbf{\underline{question}} to create large-scale counterfactual visual question pairs. Since gender counterfactuals can be introduced to either the \textbf{image} or \textbf{question} components, we can control the gender counterfactual modality to support bias evaluation in both multimodal and unimodal contexts.

\ding{182} \emph{\textbf{Multimodal bias}} encompasses gender modifications across image and text. Given $occ$ with its counterfactual image pairs, alongside its stereotyped occupation $occ_{s}$, the \textbf{question} is designed to be straightforward: ``What is the $\{{gender}\}$'s occupation in this image?\textbackslash{n}'', where $\{gender\}$ is the perceived gender of \textbf{image}. The \textbf{option} is structured as: ``Options: (A) $\{occ\}$ (B) $\{occ_{s}\}$\textbackslash{n}''. 
We repeat this process for all identified stereotyped occupation pairs to develop \toolns.

\ding{183} \emph{\textbf{Visual unimodal bias}} exclusively involves gender modifications within the image. Thus, we merely repeat the construction process in \emph{Multimodal bias}, substituting the \textbf{question} template with ``What is the \textbf{person}'s occupation in this image?\textbackslash{n}'', while keeping all other steps unchanged.

\ding{184} \emph{\textbf{Language unimodal bias}} exclusively involves gender modifications in the text, yet gender presentation in images could impede this evaluation. We utilize gender-bleached \cite{van2018bleaching} images to address this. Specifically, we employ the Segment Anything model \cite{kirillov2023segment} to obtain person masks in base images, conditioning with person bounding boxes detected by Grounding DINO. Then, we grayscale the masked portion to generate gender-bleached images, which replace the corresponding base and counterfactual \textbf{image}s in \emph{Multimodal bias}, thereby enabling language bias evaluation.

\subsection{Evaluation Protocol}

\textbf{Database}. To achieve a trade-off between test significance and accessibility, we limit the number of counterfactual visual question pairs to 20 for each subject ``$\{gender\}$ $\{occ\}$''. Therefore, we develop \toolns, comprising 34,581 visual question counterfactuals across 486 occupation pairs, covering 177 occupations (85 male-dominated and 92 female-dominated) and 5,924 counterfactual image pairs, supporting bias evaluation in both multimodal and unimodal contexts.

\textbf{LVLMs}. Based on \toolns, we benchmark 15 commonly used open-source LVLMs: InstructBLIP \cite{dai2024instructblip}, LLaVA series (LLaVA1.5-7B, LLaVA1.5-13B, LLaVA1.6-13B) \cite{liu2023improved},  Minigpt-v2 \cite{chen2023minigpt}, mPLUG-Owl2 \cite{ye2023mplug}, Qwen-VL \cite{bai2023qwen}, LLaMA-Adapter-v2 (LAv2) \cite{gao2023llama}, LAMM \cite{yin2024lamm}, Otter \cite{li2023mimic}, Kosmos-2 \cite{peng2023kosmos}, Shikra \cite{chen2023shikra}, InternLM-Xcomposer2 (InternLM-XC2) \cite{dong2024internlm}, LLaVA-RLHF \cite{sun2023aligning}, and RLHF-V \cite{yu2023rlhf}. We evaluate these LVLMs using the single-turn Perplexity (PPL) \cite{klein2017opennmt} inferencer provided in \cite{shi2023chef,shi2024assessment}, which confines their output to options and computes the probability for each option. We also evaluate state-of-the-art commercial LVLMs (\ie, GPT-4o \cite{gpt4o} and Gemini-Pro \cite{team2023gemini}), obtaining answers through their official APIs.

\textbf{Metrics}. An ideal LVLM should excel in cognitive abilities while avoiding biases. Thus, given a stereotyped occupation pair $(occ_{m}, occ_{f})$ and its two visual question counterfactuals sets $Q_{m}$ (images are $occ_{m}$) and $Q_{f}$, we define the metrics for accuracy, bias, and an idealized score for this stereotyped pair. Additionally, we explain how to calculate these metrics for the entire dataset.

\ding{182} \textbf{Accuracy}. We define the accuracy $Acc$ of an occupation pair as the average percentage of correctly inferred base visual questions in $Q_{m}$ and $Q_{f}$. Overall $Acc$ of a dataset is the average $Acc$ of all pairs.

\ding{183} \textbf{Bias}. \tool supports measuring bias in two ways: probability difference (under counterfactual fairness \cite{kusner2017counterfactual}) and outcome difference (under causal fairness \cite{galhotra2017fairness}). The former examines changes in prediction probabilities of open-source LVLMs, which we detail here. The latter considers only outcomes, making it applicable to APIs, and is detailed in the Appendix. For bias based on probability difference, we first define the bias of $occ_{m}$ in an occupation pair:  
\begin{equation}
bias({occ_{m}}) = \mathbb{E}_{(q_{g}, q_{g}^{'}) \sim Q_{m}}[\mathbbm{1}_{g=g_0} (P(occ_{m}|q_{g})-P(occ_{m}|q_{g}^{'})) - \mathbbm{1}_{g=g_1} (P(occ_{m}|q_{g})-P(occ_{m}|q_{g}^{'}))],
\end{equation} 
where $q_{g}$ and $q_{g}^{'}$ are the base visual question and its counterfactual respectively, $g$ is the gender of base visual question $q_{g}$, $P(occ_{m}|q_{g})$ is the probability of choosing $occ_{m}$ under $q_{g}$, $g_0$ and $g_1$ are predefined gender groups, and $\mathbbm{1}$ is the indicator function ($\mathbbm{1}_{A}=1$ if and only if the event "A" is true).
Here We set $g_0$ as male and $g_1$ as female; thus, a positive value indicates a male bias for $occ_{m}$, while a negative value indicates a female bias. Then, we can define the bias $B_{pair}$ of an occupation pair as: 
\begin{equation}
B_{pair}(occ_{m},occ_{f}) = 0.5 \times [bias({occ_{m}}) - bias({occ_{f}})].
\end{equation} 
This represents the average probability change introduced by gender counterfactuals, where a positive value means LVLMs perceive $occ_{m}$ as more masculine than $occ_{f}$, and a negative value means LVLMs perceive $occ_{m}$ as more feminine than $occ_{f}$. For the entire dataset, we define overall bias $B_{ovl}$ as the average absolute $\vert B_{pair} \vert$ of all pairs, and $B_{max}$ as the maximum absolute $\vert B_{pair} \vert$ among all pairs. Specifically, we define micro bias $B_{micro}$ to facilitate analysis on occupation level: 
$
B_{micro}(occ_{m}) = \frac{1}{|S_{occ_{m}}|} \sum\nolimits_{i=1}^{S_{occ_{m}}}bias(occ_{m}),
$
where $S_{occ_{m}}$ is the set of pairs containing $occ_{m}$. For bias based on outcome difference, we similarly define $B^{o}_{pair}$, $B^{o}_{ovl}$, and $B^{o}_{max}$; details in Appendix.

\ding{184} \textbf{Idealized score}. To combine accuracy and bias results, we introduce an idealized paired stereotype bias test score of an occupation pair as $Ipss = Acc \times (1 - \vert B_{pair} \vert)$. We define the overall $Ipss$ of the entire dataset as the average $Ipss$ of all pairs. Similarly, we denote the idealized score for bias based on outcome difference as $Ipss^{o}$.

Under the above definitions, an ideal LVLM is characterized by high accuracy and low absolute bias, thereby achieving a high idealized score. Besides, we conduct an option-swapping test across the dataset as LVLMs may favor certain option orders \cite{pezeshkpour2023large}. We utilize the absolute accuracy difference ($\Delta Acc$) to measure susceptibility to option order and report the average values for these metrics.

\section{Benchmark Evaluation Results}
In this section, we report the primary evaluation results of LVLMs within our \tool benchmark and summarize the key findings to understand the biases presented by the LVLMs.



\begin{table}[!t]
\caption{Results (in \%) under multimodal bias (VL-Bias), visual unimodal bias (V-Bias), and language unimodal bias (L-Bias) evaluation. \textbf{Bold} denotes the best-performing entry, while the worst performer is {\ul underlined}.}

\label{tab:overall_bias}
\resizebox{\columnwidth}{!}{
\begin{tabular}{@{}c|rrrrr|rrrrr|rrrrr@{}}
\toprule
\multirow{2}{*}{Model} & \multicolumn{5}{c|}{VL-Bias} & \multicolumn{5}{c|}{V-Bias} & \multicolumn{5}{c}{L-Bias} \\ \cmidrule(l){2-16} 
 & $Ipss\textcolor{red}{\uparrow}$ & $B_{ovl}\textcolor{blue}{\downarrow}$ & $B_{max}\textcolor{blue}{\downarrow}$ & $Acc\textcolor{red}{\uparrow}$ & $\Delta Acc\textcolor{blue}{\downarrow}$ & $Ipss\textcolor{red}{\uparrow}$ & $B_{ovl}\textcolor{blue}{\downarrow}$ & $B_{max}\textcolor{blue}{\downarrow}$ & $Acc\textcolor{red}{\uparrow}$ & $\Delta Acc\textcolor{blue}{\downarrow}$ & $Ipss\textcolor{red}{\uparrow}$ & $B_{ovl}\textcolor{blue}{\downarrow}$ & $B_{max}\textcolor{blue}{\downarrow}$ & $Acc\textcolor{red}{\uparrow}$ & $\Delta Acc\textcolor{blue}{\downarrow}$ \\ \midrule

LLaVA1.5-7B & 51.58 & 1.85 & 15.94 & 52.15 & {\ul 95.66} & 51.67 & 1.60 & 11.34 & 52.17 & {\ul 95.62} & 50.86 & 1.25 & 12.08 & 51.27 & {\ul 97.43} \\ 
LLaVA1.5-13B & 57.92 & 2.91 & 18.60 & 59.08 & 81.08 & 58.85 & 2.55 & 14.44 & 59.90 & 79.27 & 55.86 & 1.65 & 14.60 & 56.41 & 86.85 \\
LLaVA1.6-13B & 65.64 & 3.29 & 21.06 & 67.70 & 59.77 & 66.65 & 3.36 & 17.55 & 68.79 & 56.72 & 62.52 & 2.37 & 17.35 & 63.93 & 69.94 \\
MiniGPT-v2 & 58.20 & 2.72 & 16.48 & 59.74 & 73.02 & 55.30 & 1.58 & 7.43 & 56.14 & 83.97 & 54.84 & 2.05 & 13.48 & 55.95 & 84.63 \\
mPLUG-Owl2 & 72.59 & 6.48 & 34.02 & 77.56 & \textbf{8.84} & 73.26 & 5.77 & 31.50 & 77.68 & 9.07 & 70.37 & 4.75 & 22.58 & 73.92 & 11.45 \\
LAv2 & 55.31 & 0.60 & 7.38 & 55.67 & 86.16 & 55.16 & 0.42 & 6.78 & 55.40 & 86.39 & 51.72 & 0.34 & 2.22 & 51.91 & 95.45 \\
InstructBLIP & \textbf{74.26} & 4.10 & 19.94 & 77.52 & 14.05 & \textbf{75.06} & 3.23 & 18.02 & 77.61 & 13.60 & \textbf{71.83} & 3.41 & 16.94 & 74.42 & 19.54 \\
Otter & 62.68 & 1.82 & 9.25 & 63.96 & 59.11 & 62.54 & 1.48 & 8.46 & 63.56 & 60.38 & 59.71 & 0.93 & 4.65 & 60.36 & 68.99 \\
LAMM & 54.51 & 1.63 & 10.09 & 55.24 & 85.69 & 57.54 & 0.62 & 4.33 & 57.94 & 77.85 & 56.13 & 0.91 & 3.72 & 56.67 & 80.50 \\
Kosmos-2 & {\ul 48.96} & \textbf{0.22} & \textbf{0.93} & {\ul 49.58} & 70.66 & {\ul 48.95} & \textbf{0.21} & \textbf{0.95} & {\ul 49.53} & 72.69 & {\ul 49.94} & \textbf{0.03} & \textbf{0.14} & {\ul 49.99} & 74.55 \\
Qwen-VL & 71.29 & 4.07 & 30.14 & 74.27 & 23.27 & 71.07 & 4.54 & 29.88 & 74.36 & 23.99 & 70.18 & 2.96 & 19.94 & 72.35 & 18.48 \\
InternLM-XC2 & 72.93 & 6.30 & {\ul 37.32} & \textbf{77.77} & 9.45 & 72.53 & 7.24 & {\ul 37.80} & \textbf{78.05} & \textbf{8.09} & \textbf{71.83} & 5.38 & {\ul 37.23} & \textbf{75.80} & \textbf{9.23} \\
Shikra & 61.08 & 3.40 & 21.56 & 63.44 & 54.48 & 60.23 & 2.10 & 14.40 & 61.66 & 63.15 & 59.69 & 3.25 & 13.86 & 61.80 & 56.65 \\
LLaVA-RLHF & 61.05 & 4.15 & 27.57 & 63.04 & 71.24 & 62.50 & 3.01 & 14.36 & 64.00 & 68.89 & 59.70 & 3.61 & 34.59 & 61.34 & 75.23 \\
RLHF-V & 67.16 & {\ul 6.96} & 27.69 & 72.34 & 15.09 & 63.83 & {\ul 10.46} & 33.05 & 71.30 & 19.02 & 64.08 & {\ul 7.36} & 33.69 & 69.25 & 27.68 \\ \midrule

RANDOM & 50.00 & 0 & - & 50.00 & 0 & 50.00 & 0 & - & 50.00 & 0 & 50.00 & 0 & - & 50.00 & 0\\
\bottomrule
\end{tabular}
}
\end{table}

\subsection{Bias Evaluation on Open-source LVLMs}

\begin{figure}[t]
    \centering
    \includegraphics[width=\textwidth]{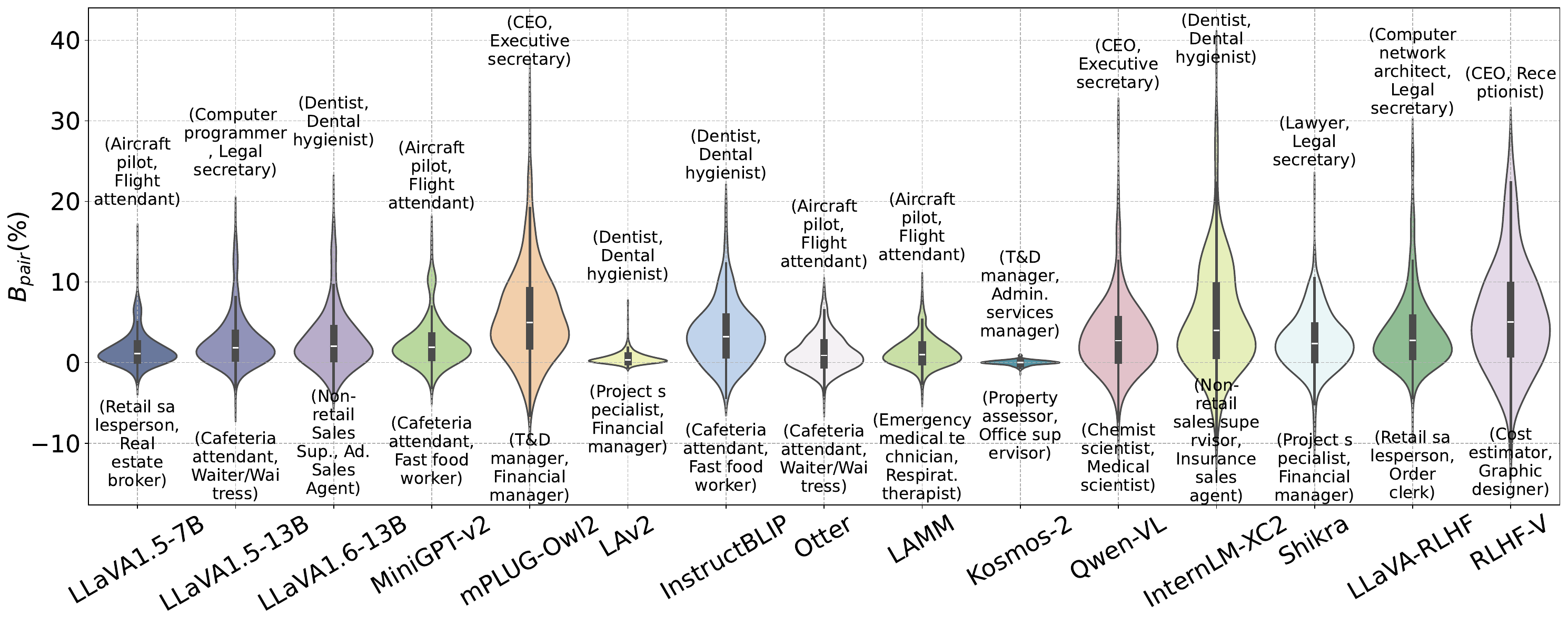}
    \caption{Distribution of LVLMs' bias ($B_{pair}$) under VL-Bias evaluation. Pairs exhibiting maximum biases (positive and negative) are plotted with occupation names. Some occupations are abbreviated.}
    \label{fig:distribution_vl_bias}
\end{figure}

We first report the overall bias evaluation results on 15 open-source LVLMs, as detailed in \cref{tab:overall_bias}. Besides 15 LVLMs, We define a RANDOM model baseline that randomly selects options, resulting in 50 $Acc$, 0 $B_{ovl}$, and 50 $Ipss$.
\cref{fig:distribution_vl_bias} presents a violin plot of the $B_{pair}$ distribution for 15 LVLMs under VL-Bias evaluation. 
From the results, we can identify: 

\ding{182} Bias is widespread among the evaluated LVLMs. 
LVLMs show an average $B_{ovl}$ of 3.09\% across different evaluation settings, indicating that the predicted probability of the correct occupation changes by this amount when the gender in the visual question is counterfactually altered. Among the Top-50 most biased occupation pairs for each LVLM, this value increases to 9.25\%. Specifically, InternLM-XC2 and RLHF-V showcase the most severe social bias, with $B_{max}$ exceeding 30\% on average across evaluation contexts. Interestingly, we observe commonalities in highly biased occupation pairs across LVLMs, such as \texttt{aircraft pilot} and \texttt{flight attendant}.

\ding{183} As LVLMs become powerful (\ie, higher $Acc$), they also become increasingly biased. This trend is likely inevitable since higher performance requires better fitting to real-world corpus, which inherently contains social biases. 
Comparing LLaVA series and LLaVA-RLHF, we conjecture three factors that increase LVLM bias: \ding{192} larger LLM models themselves may embed more bias (LLaVA1.5-7B vs. LLaVA1.5-13B); \ding{193} a larger training corpus may contain more human stereotypes (LLaVA1.5-13B vs. LLaVA1.6-13B); \ding{194} Reinforcement Learning from Human Feedback \cite{ouyang2022training} may inadvertently introduce human preferences (LLaVA1.5-13B vs. LLaVA-RLHF). While these factors boost performance, they also heighten bias risk, necessitating a balanced approach for responsible LVLMs.


Besides the primary observations, we also notice that \ding{182} For $Ipss$, almost all LVLMs (except Kosmos-2) surpass the RANDOM model, with InstructBLIP achieving the highest score (73.72\% across contexts) by effectively balancing accuracy and bias. \ding{183} Regarding occupation cognitive ability, LVLMs exhibit inconsistently. Kosmos-2 performs worse than the RANDOM model, primarily because its specialized training on detection datasets hinders its ability to understand the provided options. Additionally, all LVLMs are susceptible to option order, with LLaVA1.5 and LAv2 experiencing the most severe effects, averaging 96.24\% and 89.33\% $\Delta Acc$ across contexts, respectively.

\subsection{Case Studies on Commercial LVLMs}
Besides open-source LVLMs, we evaluate the gender bias of state-of-the-art black-box commercial LVLM APIs, including GPT-4o \cite{gpt4o} and Gemini-Pro \cite{team2023gemini}. Due to query limitations and costs, we restrict the evaluation dataset to the Top-10 biased occupation pairs listed in \cref{tab:top10_occupation_pairs}. \cref{fig:case_study} presents the average results (bias based on outcome difference) across three evaluation contexts.
Despite their widespread acclaim for outstanding performance, \textbf{GPT-4o and Gemini-Pro also exhibit biases similar to those seen in open-source LVLMs}, highlighting the pervasive bias risk inherent in LVLMs. Specifically, Gemini-Pro faces significant bias issues, exhibiting nearly \textbf{3 times} the bias level of GPT-4o. In contrast, GPT-4o achieves notably low $B^{o}_{ovl}$, resulting in the highest $Ipss^{o}$ and outperforming InstructBLIP (the leading open-source model).

\begin{minipage}[!t]{\textwidth}
    \begin{minipage}[ht]{0.47\textwidth}
        \centering
        \includegraphics[width=1\textwidth]{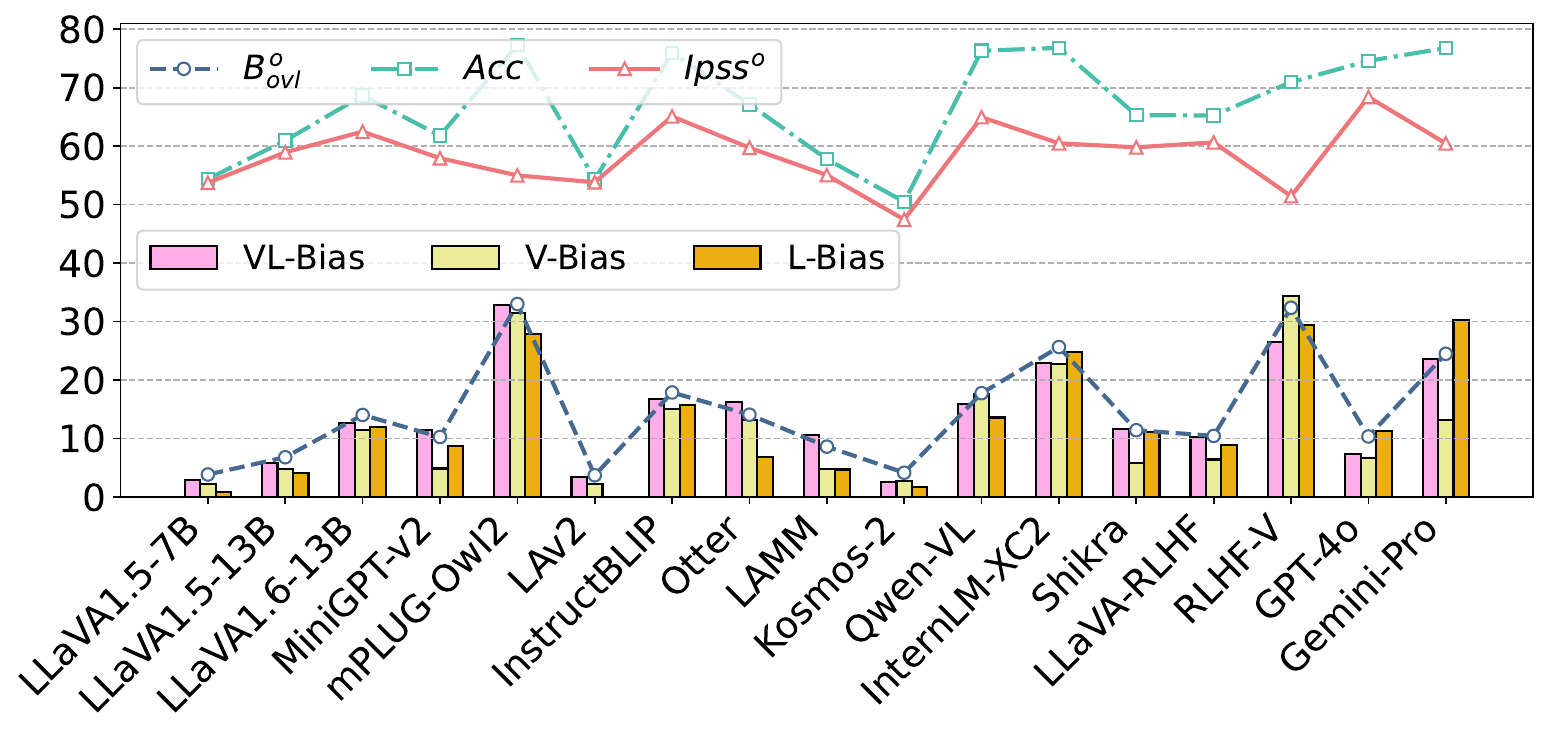}
        \makeatletter\def\@captype{figure}\makeatother
        \caption{Results (in \%) of commercial LVLMs on Top-10 biased pairs. The polylines denote the average results across evaluation contexts.}
        \label{fig:case_study}
    \end{minipage}
        \hspace{3mm}%
    \begin{minipage}[ht]{0.47\textwidth}
        \centering
        \resizebox{0.84\textwidth}{!}{
        \begin{tabular}{@{}lrrr@{}}
            \toprule
            Occupation Pair ($occ_m, occ_f$) & $B_{pair}$ & $Acc$ & $Ipss$ \\ \midrule
            Dentist, Dental hygienist & 37.32 & 70.62 & 44.14 \\
            CEO, Executive secretary & 34.02 & 78.12 & 51.59 \\
            Surgeon, Surgical technologist & 30.59 & 57.52 & 39.94 \\
            Lawyer, Legal secretary & 30.19 & 52.63 & 36.83 \\ \midrule
            \multirow{2}{*}{\begin{tabular}[c]{@{}l@{}}Refractory machinery mechanic, \\ Filling machine operator\end{tabular}} & \multirow{2}{*}{26.92} & \multirow{2}{*}{74.06} & \multirow{2}{*}{54.03} \\
             & & & \\ \midrule
            Aircraft pilot, Flight attendant & 25.41 & 86.70 & 67.48 \\
            Com. Sys. Manager, Receptionist & 24.10 & 79.73 & 60.57 \\
            EMT, Licensed practical nurse & 23.63 & 75.17 & 57.38 \\
            Network Architect, Billing Clerk & 22.92 & 75.49 & 58.74 \\
            Financial analyst, HR manager & 13.88 & 80.00 & 69.14 \\
            \bottomrule
        \end{tabular}%
        }
        \makeatletter\def\@captype{table}\makeatother\caption{Top-10 pairs with the most biased results. Some occupations are abbreviated.}
        \label{tab:top10_occupation_pairs}   
    \end{minipage}

\end{minipage}

\subsection{Bias Characteristics Analysis}
After reporting the overall bias performance, we then further study where and how these biases emerge. In particular, we focus our discussion on 14 of the 15 LVLMs (excluding Kosmos-2 as it performs worse than random choosing) evaluated under VL-Bias, with similar results for other bias contexts in the Appendix.

\textbf{Top Biased Pairs.} From the main evaluation results, we identify the Top-10 biased occupation pairs (excluding duplicated occupations). As shown in \cref{tab:top10_occupation_pairs}, in addition to the commonly examined occupation pairs that embody gender stereotypes, such as \texttt{CEO} and \texttt{executive secretary}, we also find rarely mentioned pairs like \texttt{refractory machinery mechanic} and \texttt{filling machine operator}, as well as \texttt{computer system manager} and \texttt{receptionist}, which also present severe gender bias. It suggests that bias extends across a broader range of occupations than previously documented, demonstrating the effectiveness of our benchmark in detecting bias in the workplace.

\textbf{Bias Patterns.} Among occupation pairs with high levels of bias, we identify the following patterns that are more likely to reveal bias in LVLMs. \emph{Male-dominated} occupations: management, business, finance, professional, or physically demanding categories; \emph{female-dominated} occupations: service, office and administrative support, or patient-requiring categories. These combination patterns reflect societal stereotypes about gender roles in occupations, and the resulting biased behavior in LVLMs, in turn, indicates their absorption of such stereotypes during the training process. Interestingly, at the intersection of finance and management categories, LVLMs may encode stronger male stereotypes for financial roles. For example, InstructBLIP consistently shows positive biases towards pairs consisting of the male-dominated \texttt{financial analyst} and four female-dominated manager occupations (excluding \texttt{financial manager}). Such phenomena are also observed in other LVLMs.


\begin{figure}[h]
    \centering
    \subfigure[Connection between VL-Bias and Labor Statistics]{
        \includegraphics[width=0.46\textwidth]
        {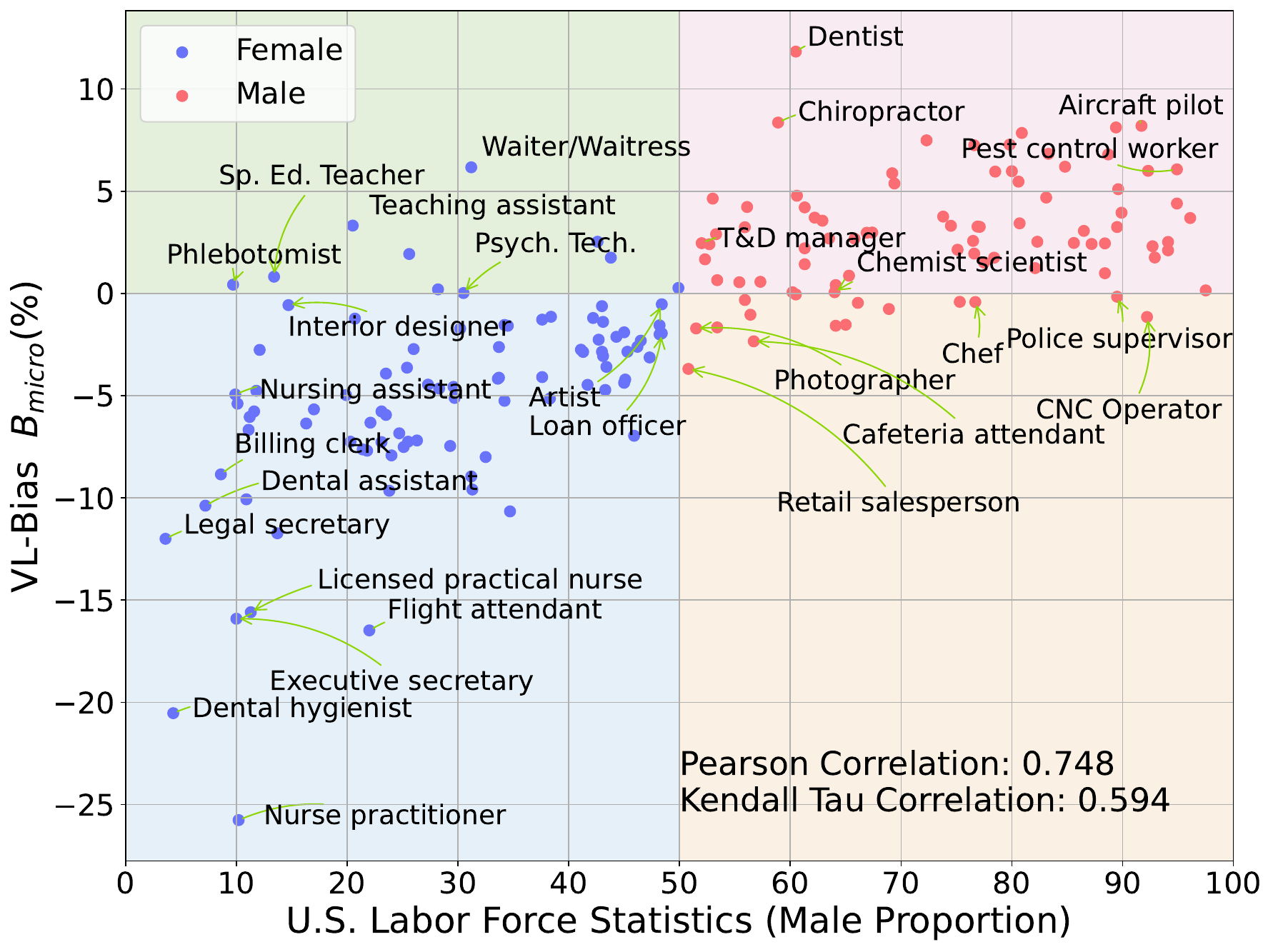}
        \label{fig:bias_vs_labor}
    }
    \subfigure[Relationship between V-Bias and L-Bias]{
        \includegraphics[width=0.46\textwidth]
        {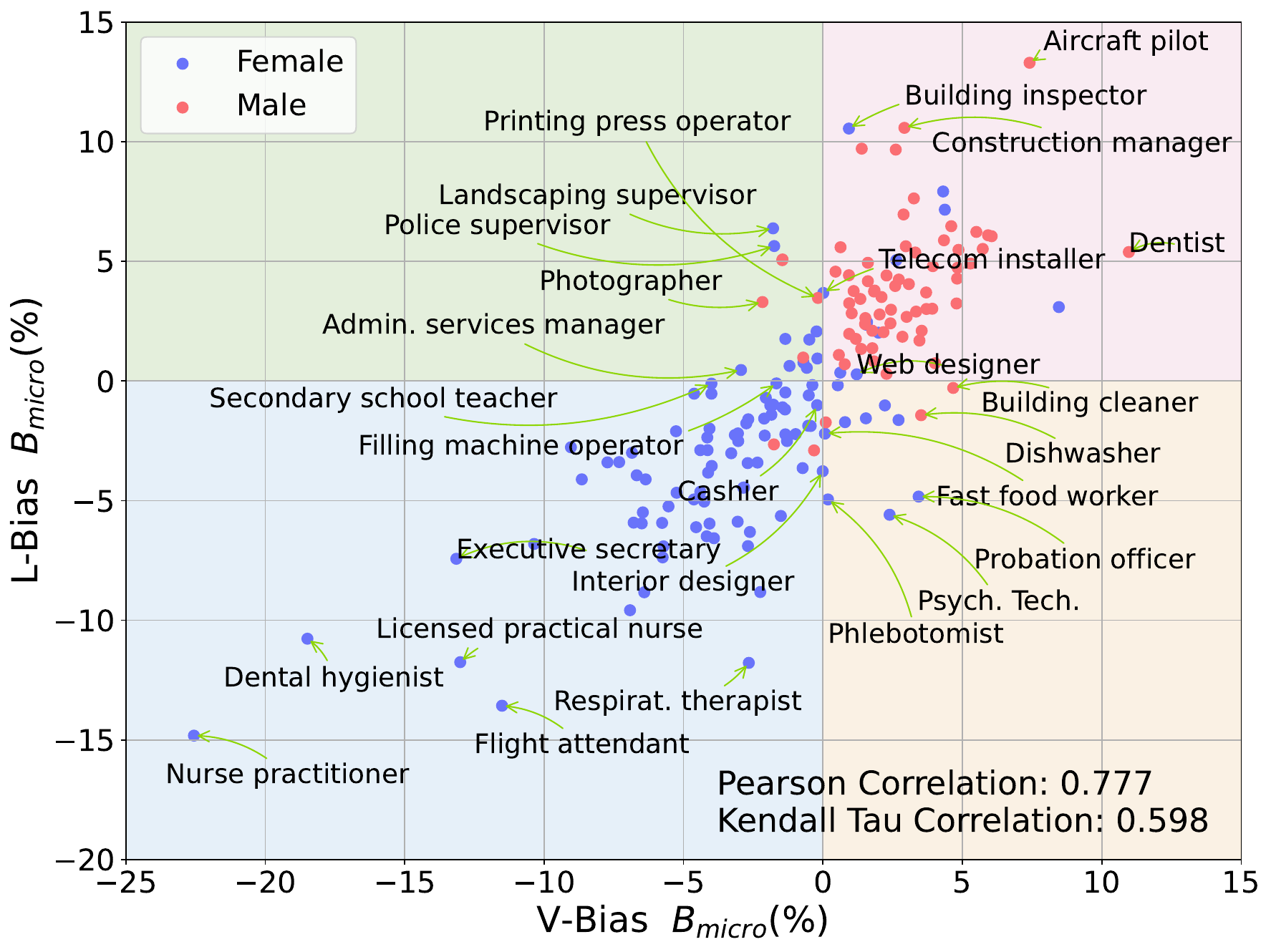}
        \label{fig:visual_bias_vs_language_bias}
    }
    \caption{Both results show occupation-level bias $B_{micro}$ of InstructBLIP. Scatter plots are divided into four colored quadrants, and points are colored to indicate male- or female-dominated occupations. \textbf{(a)} Occupation bias generally aligns with U.S. Labor Force Statistics. \textbf{(b)} Bias exhibits strong consistency in visual and language modal. \textbf{\textit{Similar results for other LVLMs are in the Appendix.}}}
\end{figure}

\textbf{Bias Direction.} Here, we further examine the gender bias direction (\ie, the sign of $B_{pair}$). A positive $B_{pair}$ indicates that LVLMs perceive male-dominated occupations as more masculine and female-dominated occupations as more feminine, while a negative $B_{pair}$ indicates the opposite. As shown in \cref{fig:distribution_vl_bias}, LVLMs exhibit positive biases for most occupation pairs, indicating that their biases mirror real-world stereotypes. However, we also observe that a small subset of occupation pairs exhibit negative bias. For example, among 10 occupation pairs involving female-dominated \texttt{financial manager}, InstructBLIP presents negative biases in eight pairs, indicating that \texttt{financial manager} is generally perceived as more masculine. The remaining two positive pairs involve industry/engineering roles, suggesting these are seen as even more masculine.


\textbf{Observed Bias vs. Real-world Labor Statistics.} To further understand the observed bias and its connection to real-world statistics, we compare occupation-level biases ($B_{micro}$) with U.S. Labor Force Statistics \cite{USLabor}, as depicted in \cref{fig:bias_vs_labor} for InstructBLIP (others shown in Appendix). The scatter plot is divided into four colored quadrants, where LVLMs' bias aligns with labor force data in quadrants one and three, but opposes it in quadrants two and four. We observe that bias presented by LVLMs generally aligns with labor statistics, as most occupations fall into quadrants one and three. Additionally, we measure the correlation between LVLM bias and occupation proportions using the Pearson Correlation ($\rho$) and Kendall-Tau Rank Correlation ($\tau$). All LVLMs exhibit strong correlations, with average $\rho$ and $\tau$ coefficients of 0.68 and 0.56, respectively, underscoring that LVLM biases inherit real-world skews.

\subsection{Relationship between Visual and Language Bias}

\textbf{V-Bias vs. L-Bias}. LVLM's inference involves complex cross-modal interactions between vision and language. To characterize their bias relationship, we use scatter plots comparing occupation-level bias ($B_{micro}$) in visual modal (V-Bias, $x$-axis) and language modal (L-Bias, $y$-axis), as shown in \cref{fig:visual_bias_vs_language_bias} for InstructBLIP (others shown in Appendix). Among the LVLMs (excluding Kosmos-2), we observe strong alignment between the direction of visual modal bias and language modal bias, with occupations clustering in quadrants one and three. Besides, there is a strong correlation in the extent of these biases, with average $\rho$ and $\tau$ coefficients of 0.77 and 0.60, respectively.
The consistency of bias across modalities in LVLMs likely stems from their LLM-centric nature. LVLMs' modality interfaces are trained via LLM-centric alignment to project multimodal information into an LLM-understandable space. Then, LLM conducts the reasoning process based on the fused information, leading to consistent biases in the two modalities.

\begin{wraptable}{hr}{0.45\textwidth}
\caption{Results (in \%) of CLIP models and part of corresponding LVLMs under V-Bias. CLIP models are highlighted in \sethlcolor{gray!20}\hl{gray}.}
\label{tab:clip_bias} 
\resizebox{0.45\textwidth}{!}{
            \begin{tabular}{@{}crrrrrr@{}}
                \toprule
                Model & $Ipss^{o}\textcolor{red}{\uparrow}$ & $B^{o}_{ovl}\textcolor{blue}{\downarrow}$ & $B^{o}_{max}\textcolor{blue}{\downarrow}$ & $Acc\textcolor{red}{\uparrow}$ & $\rho$ & $\tau$ \\ \midrule
                \rowcolor[gray]{0.8} CLIP ViT-L \cite{radford2021learning} & 72.69 & 9.11 & 50.00 & 79.49 & - & - \\  
                LLaVA1.5-7B & 52.01 & 0.85 & 18.38 & 52.17 & 0.00 & 0.01 \\
                LLaVA1.5-13B & 59.00 & 2.42 & 16.04 & 59.90 & 0.10 & 0.10 \\
                LLaVA1.6-13B & 66.30 & 4.50 & 25.61 & 68.79 & 0.41 & 0.29 \\
                LLaVA-RLHF & 62.47 & 3.25 & 21.02 & 64.00 & 0.21 & 0.14 \\
                InternLM-XC2 & 70.56 & 9.94 & 56.88 & 78.05 & 0.52 & 0.35 \\ \midrule
                \rowcolor[gray]{0.8} EVA-G \cite{sun2023eva} & 73.45 & 8.12 & 60.00 & 79.67 & - & -  \\  
                MiniGPT-v2 & 55.45 & 1.96 & 18.75 & 56.14 & 0.19 & 0.10 \\
                InstructBLIP & 73.33 & 5.66 & 40.25 & 77.61 & 0.44 & 0.33 \\
                \bottomrule
            \end{tabular}%
        }
\end{wraptable}

\textbf{V-Bias correlation between CLIP and LVLMs}. Moreover, we evaluate CLIP models involved in LVLMs under V-Bias to see whether visual biases in CLIP transfer to LVLMs. \cref{tab:clip_bias} reports the average results (bias based on outcome difference), revealing that CLIP models exhibit even greater biases than LVLMs. However, biases at the occupation pair level between CLIP and LVLMs show relatively weak correlations (0.26 and 0.19 for $\rho$ and $\tau$ coefficients on average). Further analysis of each counterfactual visual question pair reveals subtle differences in the biases of CLIP and LVLMs. For instance, there are only 874 (11\%) overlapping biased samples between EVA-G \cite{sun2023eva} and InstructBLIP, indicating that LVLMs display distinct behaviors from CLIP models, despite both sharing the same image encoder. The result also indirectly supports our analysis that the consistent bias across visual and language modalities stems from the LLM-centric nature.

\section{Conclusion and Future Work}
\label{sec:conclusion}
This paper introduces \tool benchmark, the first to evaluate occupation-related gender bias in LVLMs under individual fairness criteria. \tool comprises 34,581 visual question counterfactual pairs covering 177 occupations, supporting bias probing in both multimodal and unimodal contexts. The extensive evaluation of 15 open-source LVLMs and the most advanced commercial LVLMs (GPT-4o and Gemini-Pro) reveals pervasive biases within these models. As LVLM development races forward, \tool offers a vital benchmark to assess social bias risks before their real-world deployment.

\textbf{Limitation.} Despite our best efforts, we acknowledge three major limitations in our research. First, the generation pipeline may contain latent biases, which may contribute to implicit biases exhibited by LVLMs. Therefore, the findings we present to enhance the understanding of bias in LVLMs should be interpreted within the context of our experiments.
Second, our benchmark is confined to a binary gender framework, whereas gender is inherently fluid, as discussed in Sec. \ref{sec:terminology}. Future research should incorporate a broader range of gender identities, including those from the LGBTQIA+ community, to provide a more comprehensive scope of bias assessment in LVLMs. Third, since our benchmark is based on U.S. Labor Force Statistics, the bias analysis primarily focuses on the Western world. It is crucial to extend this analysis to encompass diverse cultures and countries in future work.

\textbf{Ethical statement and broader impact}. 
Our paper aims to probe and benchmark the occupation-related gender bias in LVLMs. While evaluation results may raise ethical concerns and potentially harm readers, our intention is not to cause harm. Rather, our work seeks to facilitate bias evaluation for LVLMs, serving as a crucial initial step toward mitigating discriminatory outcomes. 
\medskip
{
\small
\bibliographystyle{unsrt}
\bibliography{main}
}


\end{document}